\documentclass[sigconf]{acmart}

\usepackage{xcolor}

\usepackage{booktabs}       
\usepackage{amsfonts}       
\usepackage{xurl}
\usepackage{array}

\usepackage{amsmath}
\usepackage{color, colortbl}
\usepackage{amsthm}
\usepackage[linesnumbered,ruled,vlined]{algorithm2e}

\usepackage{comment}
\usepackage{mathtools}

\usepackage{microtype}
\usepackage{graphicx}

\usepackage{caption}
\usepackage{subcaption}

\usepackage{xspace}
\usepackage{forloop}
\usepackage{multirow}
\usepackage{booktabs}
\usepackage{algorithm,algorithmic,refcount}
\usepackage{footnote}

\usepackage{graphicx}
\usepackage[shortlabels]{enumitem}
\usepackage{bbm}
\usepackage{float}
\usepackage[most]{tcolorbox}
\usepackage{etoolbox}
\usepackage{bm}

\usepackage[mathscr]{euscript}
\usepackage{accents}

\DeclareSymbolFont{rsfs}{U}{rsfs}{m}{n}
\DeclareSymbolFontAlphabet{\mathscrsfs}{rsfs}

\newcommand{\Expect}{{\mathbb E}}

\newcommand{\eps}{\varepsilon}

\renewcommand{\l}{\vert}

\def\bI{{\boldsymbol I}}

\def\beps{{\boldsymbol \epsilon}}

\def\bx{{\boldsymbol x}}

\def\beps{{\boldsymbol \eps}}

\def\btheta{{\boldsymbol \theta}}

\def\cD{{\mathcal D}}

\usepackage{color}

\newcommand{\norm}[1]{\left\|#1\right\|}

\newcommand{\maxx}[1]{\underset{#1}{\mathrm{max}} \ }

\definecolor{mylavender}{rgb}{0.90, 0.90, 0.98}

\AtBeginDocument{%
  \providecommand\BibTeX{{%
    \normalfont B\kern-0.5em{\scshape i\kern-0.25em b}\kern-0.8em\TeX}}}

\setcopyright{acmlicensed}
\copyrightyear{2024}
\acmYear{2024}
\acmMonth{May}
\acmDOI{XXXXXXX.XXXXXXX}

\acmConference[FL\@FM-TheWebConf'24]{International Workshop on Federated Foundation Models for the Web 2024 }{May 14, 2024}{Singapore}

\begin{document}

\title{Stable Diffusion-based Data Augmentation for Federated Learning with Non-IID Data}

\author{Mahdi Morafah}
\authornote{Work done while doing an internship at Qualcomm AI Research. Qualcomm AI Research is an initiative of Qualcomm Technologies, Inc. and/or its subsidiaries.}
\authornote{Correspondence: mmorafah@ucsd.edu}
\affiliation{
  \institution{University of California San Diego}
  \country{USA}
}

\author{Matthias Reisser}
\affiliation{
  \institution{Qualcomm AI Research}
  \country{USA}
}

\author{Bill Lin}
\affiliation{
  \institution{University of California San Diego}
  \country{USA}
}

\author{Christos Louizos}
\affiliation{
  \institution{Qualcomm AI Research}
  \country{Netherlands}
}

\begin{abstract}
The proliferation of edge devices has brought Federated Learning (FL) to the forefront as a promising paradigm for decentralized and collaborative model training while preserving the privacy of clients' data. However, FL struggles with a significant performance reduction and poor convergence when confronted with Non-Independent and Identically Distributed (Non-IID) data distributions among participating clients.
While previous efforts, such as client drift mitigation and advanced server-side model fusion techniques, have shown some success in addressing this challenge, they often overlook the root cause of the performance reduction—the absence of identical data accurately mirroring the global data distribution among clients. In this paper, we introduce \textbf{``Gen-FedSD''} a novel approach that harnesses the powerful capability of state-of-the-art text-to-image foundation models to bridge the significant Non-IID performance gaps in FL. In Gen-FedSD, each client constructs textual prompts for each class label and leverages an of-the-shelf state-of-the-art pre-trained Stable Diffusion model to synthesize high-quality data samples. The generated synthetic data is tailored to each client's unique local data gaps and distribution disparities, effectively making the final augmented local data IID. Through extensive experimentation, we demonstrate that Gen-FedSD achieves state-of-the-art performance and significant communication cost savings across various datasets and Non-IID settings.
\end{abstract}


\maketitle

\section{Introduction}

Federated Learning (FL) has emerged as a promising framework enabling decentralized training of machine learning models. It allows, under the orchestration of a central server, for a massive amount of edge devices to collaborate in training the model, while keeping the client's data local~\cite{mcmahan2017communication, kairouz2021advances}. However, FL struggles with a significant performance degradation and slow convergence when confronted with Non-Independent and Identically Distributed (Non-IID) data distributions among participating clients (a.k.a data heterogeneity)~\cite{zhao2018federated, li2022federated,zhu2021federated, haddadpour2019convergence}. 

Numerous studies have been conducted to address the challenge of data heterogeneity. Some studies propose modifications in the local objectives of the clients to prevent the deviation of locally trained models from the global model~\cite{zhao2018federated, karimireddy2020scaffold, li2021model}. Meanwhile, others introduce server-side model fusion techniques~\cite{lin2020ensemble, sattler2021fedaux, yurochkin2019bayesian, wang2020federated}. While prior endeavors have exhibited some enhancements in addressing the challenge of data heterogeneity, a crucial foundational issue remains unaddressed. Specifically, \textbf{clients lack access to identical data that accurately represent the global data distribution.} This is the primary factor to the substantial performance degradation from the ideal Independent and Identically Distributed (IID) data distribution in FL.

Recently, several works have proposed data augmentation techniques to construct IID data for each client's local data.~\citet{zhao2018federated} used an auxiliary public dataset to augment each client's local data. However, this can introduce an additional set of challenges; firstly, there could be an inherent domain difference between the local and public data and, secondly, gathering this public data may be extremely difficult in practice~\cite{zhu2021data, zhang2023gptfl}. Alternatively, other methods suggest synthesizing the local data using techniques such as dataset distillation~\cite{wang2018dataset} and sharing it with other clients~\cite{zhou2020distilled, yoon2021fedmix, shin2020xor, hu2022fedsynth}. Nonetheless, this approach may raise privacy concerns, and the synthesized data might not exhibit high quality, which limits its effectiveness.

In this paper, we introduce a novel \textbf{``\underline{Gen}erative \underline{Fed}erated Learning with \underline{S}table \underline{D}iffusion'' (Gen-FedSD)} , leveraging pre-trained state-of-the-art Stable Diffusion (SD)~\cite{rombach2022high} to address the Non-IID challenge in FL. In Gen-FedSD, clients construct textual prompts of the entire class labels and employ a pre-trained state-of-the-art SD model to synthesize high-quality data samples via the constructed input text prompts. The locally generated images are tailored to fill the data gaps and distribution disparities of each client, effectively aligning local data distributions with the global one, so as to emulate the IID setting.

We empirically evaluate Gen-FedSD on CIFAR-10 and CIFAR-100 datasets under various Non-IID settings. We make three observations. (1) Gen-FedSD significantly enhances accuracy by at least 12\% and 6\% on CIFAR-10 and CIFAR-100 datasets, respectively, under mild data heterogeneity. (2) As the heterogeneity levels increase, Gen-FedSD exhibits even more pronounced performance improvements. Notably, under extreme data heterogeneity, we observe accuracy improvements of at least 20\% and 7\% for the CIFAR-10 and CIFAR-100 datasets.
(3) Gen-FedSD accelerates convergence and reduces communication costs. Additionally, we investigate the impact of two different prompt designs for image generation and find that prompt diversity can further enhance the performance of Gen-FedSD.

\textbf{Organization.} The rest of the paper is organized as follows. In section~\ref{sec:related-works} we review the related works. In section~\ref{sec:background} we discuss the necessary background on diffusion models, image generation using Stable Diffusion, and federated learning. In section~\ref{sec:method} we explain our proposed method. In section~\ref{sec:experiments} we present our experimental results, and finally in section~\ref{sec:conclusion} we conclude our work and discuss future works.
\section{Related Works} \label{sec:related-works}

\textbf{Data Heterogeneity in FL.} FedAvg was first introduced by~\citet{mcmahan2017communication} and has shown to be successful in several applications~\cite{chen2019federated, ramaswamy2019federated, feng2020pmf, saputra2019energy}. Unfortunately, the performance of FedAvg drops significantly in presence of data heterogeneity due to clients' local models drifting from the global model~\cite{zhao2018federated, karimireddy2020scaffold}. Several studies attempt to mitigate the client drift issue by various techniques. FedProx~\cite{li2020federated} introduces a proximal term during local training to constrain the deviation of local model weights from the global model. FedNova~\cite{wang2020tackling} proposes a weighting strategy in the server-side averaging to reduce the bias in the solution. Scaffold~\cite{karimireddy2020scaffold} employs control variates to correct the drift in local updates, and FedOpt~\cite{reddi2020adaptive} proposed federated versions of adaptive optimizers by replacing the server-side averaging with a gradient-based server optimizer. 
Other approaches such as server-side model fusion techniques~\cite{lin2020ensemble, sattler2021fedaux, yurochkin2019bayesian, wang2020federated}, neuron matching techniques~\cite{yurochkin2019bayesian, wang2020federated, li2022federated, singh2020model} and personalization~\cite{kulkarni2020survey, deng2020adaptive, collins2021exploiting, morafah2023practical, hanzely2020lower, vahidian2023efficient} have been explored.

\textbf{Data Augmentation in FL.} Although prior works have shown some success in addressing data heterogeneity, they often overlooked the primary factor that clients do not have access to data that accurately mirrors the global data distribution. To address this, several works have proposed techniques to augment clients' local data with either an auxiliary public dataset~\cite{zhao2018federated} or synthetic data~\cite{zhou2020distilled, yoon2021fedmix, shin2020xor, hu2022fedsynth}. Unfortunately, forming an auxiliary public dataset may be extremely difficult in practice~\cite{zhu2021data, zhang2023gptfl}, and it can introduce additional heterogeneity due to its domain difference with the clients' local private data. Other works propose techniques to synthesize the clients' local private data and share it with the server or other clients~\cite{goetz2020federated, zhou2020distilled, yoon2021fedmix, shin2020xor, hu2022fedsynth}. For example,~\citet{yoon2021fedmix} proposed for a client to exchange the updated model parameters as well as its mashed (or averaged) data,~\citet{shin2020xor} proposed a XorMixup technique to encode each clients' local data and share it with the server, and in~\cite{hu2022fedsynth} each client learns and transmits a light-weight synthetic dataset to the server. However, this approach can compromise the privacy and the performance improvement is low due to the low quality of the synthesized data. Training a GAN in FL and utilizing it to synthesize data has been also explored by~\citet{jeong2018communication, zhu2021data}. However, training a GAN in FL under data heterogeneity can lead to poor performance and low quality image generation. Our work departs from the limitations of prior work by leveraging a state-of-the-art pre-trained stable diffusion model to synthesize high-quality images. Gen-FedSD circumvents challenges associated with training generative models in FL and eliminates the privacy risks of prior approaches by generating the data at the client level.

\textbf{Diffusion Models.} Recently, Diffusion Models (DM)~\cite{ho2020denoising, nichol2021improved} have gained increasing interest in the field of generative modeling. They demonstrate state-of-the-art image synthesis quality and, compared to GANs, offer greater stability during training and scalability~\cite{dhariwal2021diffusion, stypulkowski2024diffused, rombach2022high, ho2022cascaded, sauer2023adversarial}. DMs have been applied to various image processing applications, including super-resolution, image inpainting, and semantic segmentation~\cite{saharia2022image, li2022srdiff, saharia2022palette, baranchuk2021label}. 
Notably, there has been a recent development of high-resolution text-to-image DMs such as Stable Diffusion (SD)~\cite{rombach2022high}, DALL-E 2~\cite{ramesh2022hierarchical}, Imagen~\cite{saharia2022photorealistic}, and GLIDE~\cite{nichol2021glide}. Recent works have demonstrated that augmenting training data with synthetic data generated using text-to-image DMs improves zero-shot and few-shot image classification performance~\cite{he2022synthetic, trabucco2023effective}, as well as ImageNet classification accuracy and robustness~\cite{azizi2023synthetic, sariyildiz2022fake, bansal2023leaving}. 
\section{Background} \label{sec:background}

\begin{figure*}[th]
    \centering
     \includegraphics[width=\textwidth, height=0.55\textwidth]{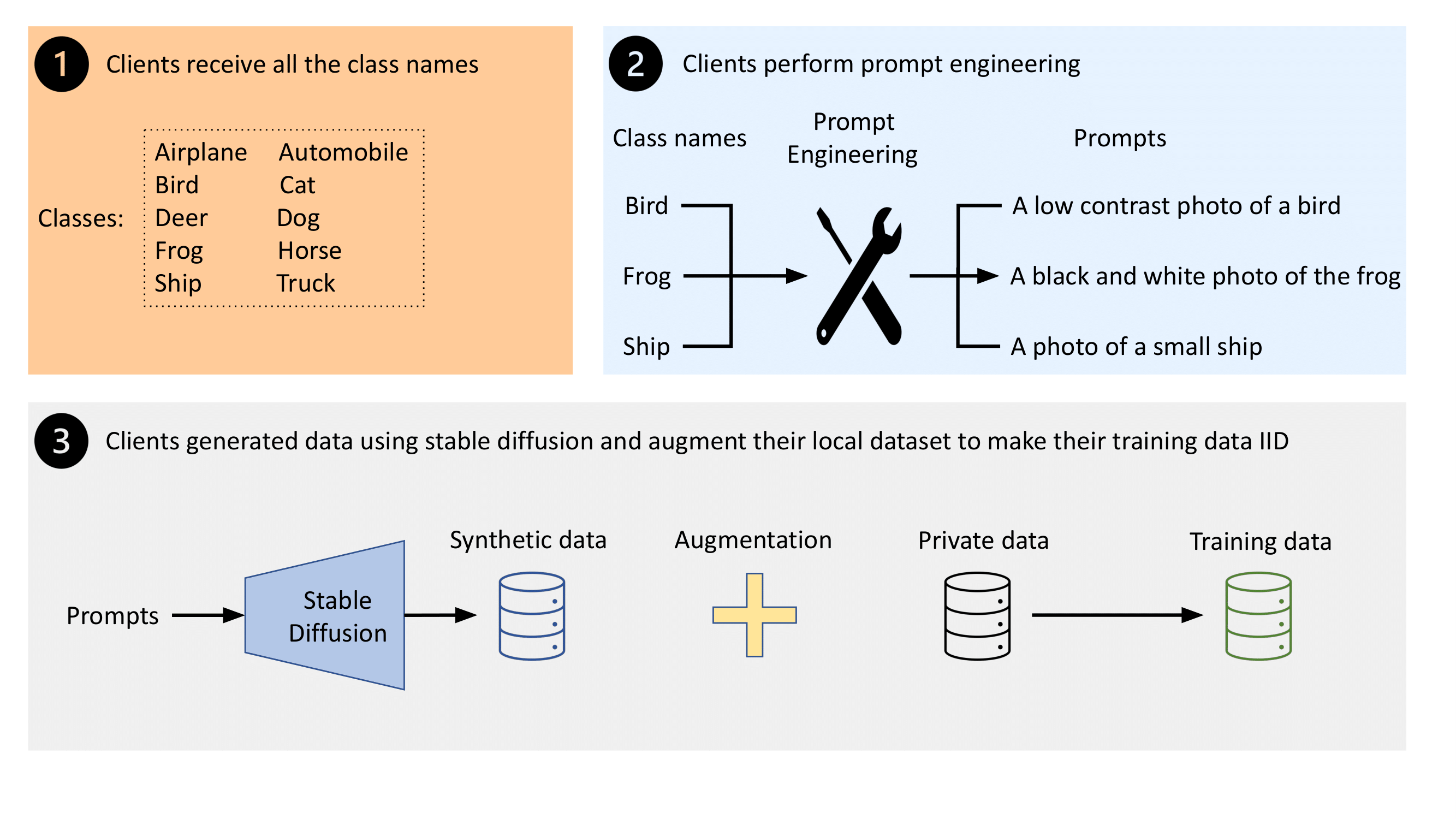}
    \caption{Overview of Gen-FedSD.}
    \label{fig1:teaser}
\end{figure*}

\subsection{Diffusion Models}
Diffusion Models (DMs)~\cite{ho2020denoising, nichol2021improved} operate by gradually transforming a clean image into pure noise through a process of forward diffusion, followed by learning to recover the original clean image from the pure noise in reverse diffusion. In the forward diffusion process, a Gaussian noise with zero mean and variance $\beta_t$ is slowly added in $T$ timesteps to degrade a clean image ($\bx_0 \sim q(\bx_0)$) and ultimately converge into pure Gaussian noise at step $T$ ($\bx_T$). The noisy image at each timestep $t$ from the previous timestep $t-1$ can be derived using the following equation:
\begin{align*}
q(\bx_t\vert \bx_{t-1}) = \mathcal{N}(\bx_t; \sqrt{1-\beta_t}\bx_{t-1}, \beta_t \bI),
\end{align*}
where $t\in[1,T]$ and $0 < \beta_{1:T} < 1$ denote the noise scale scheduling. In the reverse diffusion process, a noise predictor model $\beps_{\theta}$ is trained to predict the noise $\beps_t$ at each timestep $t$ using the following loss function:
\begin{align*}
    \mathcal{L}_{} = \Expect_{t, \bx_0, \beps}[\norm{\beps_t - \beps_{\theta}(\bx_t, t)}_2^2].
\end{align*}
For further information about DMs, we refer readers to~\cite{yang2023diffusion, cao2024survey}.

\subsection{Image Generation using Stable Diffusion}
Stable Diffusion (SD) is a type of Latent Diffusion Model (LDM)~\cite{rombach2022high} where the denoising process occurs in the latent space of robust pre-trained Variational Autoencoders (VAEs)~\cite{kingma2013auto, razavi2019generating}, as opposed to the original pixel space. Notably, SD incorporates cross-attention layers into its model architecture, providing the capability for general conditioning inputs on image generation, such as text or bounding boxes.
To generate an image using input text, SD utilizes the pre-trained CLIP text encoder~\cite{radford2021learning} to obtain text embeddings. Subsequently, these text embeddings are fed into the model and cross-attention layers, serving as a conditioning signal for image generation. In our work, each client leverages a state-of-the-art pre-trained SD to generate its missing data points, relative to the global data distribution, via input text prompts. For more detailed discussion about text-to-image DMs we refer readers to~\cite{zhang2023text}.

\subsection{Federated Learning}

In federated learning, a set of clients denoted as $S$ collaborate to train a machine learning model $f(\cdot; \btheta)$ under the orchestration of a central server. At each communication round $r$, a random fraction $C \in (0, 1]$ of clients is selected. The server then distributes the global model's parameters $\btheta_r^g$ to the selected clients $S_r$ and each client $k \in S_r$ independently updates the model based on its private dataset $\cD_k$. Subsequently, the clients transmit the updated parameters ($\{{\widehat \btheta_k}\}_{k \in S_r}$) back to the sever to update the global model. FedAvg~\cite{mcmahan2017communication}, which has been the de facto federated learning method, employs a straightforward aggregation for updating the global model, as denoted below:
\begin{align*}
    \btheta_{r+1}^g = \sum_{k \in \l S_r\l} \frac{\l \cD_k \l}{\sum_{k \in \l S_r\l} \l \cD_k \l} {\widehat \btheta_k}.
\end{align*} 
This process continues for a total of $R$ communication rounds.

\section{Gen-FedSD} \label{sec:method}
\subsection{Overview}
Figure~\ref{fig1:teaser} illustrates the overall process of Gen-FedSD, which consists of three key steps. First, the clients receive all the class labels' names from the server. Second, the clients apply prompt engineering techniques to transform the received class labels' names into text prompts. In the third step, the clients utilize a state-of-the-art pre-trained SD to synthesize custom images, addressing local data distribution disparities and gaps. Subsequently, the clients augment their local data with the generated synthetic data and initiate the federation. The following sections provide detailed explanations for each step.

\begin{figure*}[ht]
    \centering
    \begin{subfigure}[b]{0.49\textwidth}
         \centering
         \includegraphics[width=.6\textwidth]{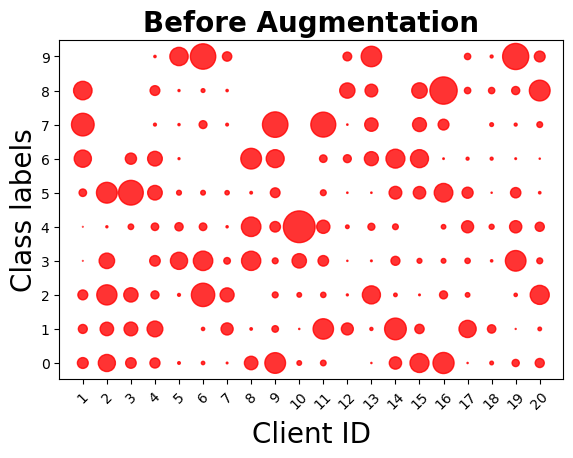}
         \caption{}
         \label{fig2:a}
     \end{subfigure}
     \begin{subfigure}[b]{0.49\textwidth}
         \centering
         \includegraphics[width=.6\textwidth]{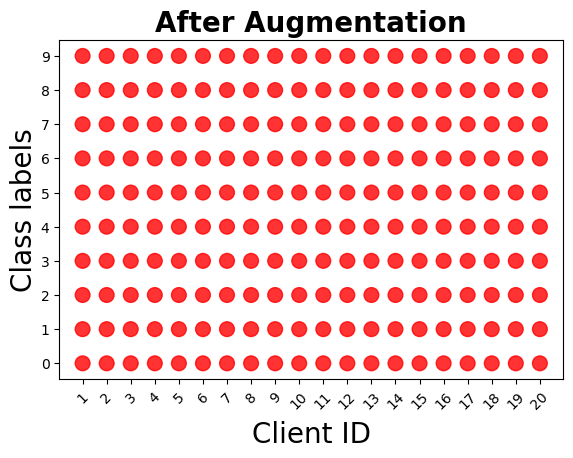}
         \caption{}
         \label{fig2:b}
     \end{subfigure}

    \caption{Illustration of clients data distribution for the CIFAR-10 dataset before and after augmentation with Gen-FedSD. In this example, 20 clients are utilized, partitioning CIFAR-10 Non-IID with a Dirichlet distribution (Dir(0.5)). Prior to augmentation, the local distributions of clients are heterogeneous (a), whereas after augmentation, the unique gaps in each client's local distribution are filled, resulting in a final locally augmented dataset that is independent and identically distributed (IID) for each client (b).} 
    \label{fig2}
\end{figure*}

\subsection{Steps 1-2: Client-level Prompt Generation}
Utilizing the received class labels' names from the server, the clients generate text prompts that describe each class. Various prompt engineering techniques exist, including the use of Large Language Models (LLMs)~\cite{zhang2023gptfl} or CLIP~\cite{radford2021learning}. Interested readers can refer to~\cite{gu2023systematic} for a more comprehensive exploration of these techniques.
Given that Stable Diffusion employs the CLIP text encoder for prompt intake, we adopt the original prompt engineering technique used for CLIP in our work. To obtain prompts, we apply natural language templates $\mathcal{M}$ proposed in CLIP to a given class label $y$. For instance, if $M(y) =$ "a photo of a $y$" where $M \subset \mathcal{M}$ and the class label is $y=\text{cat}$, the prompt would be "this is a photo of a cat."

\subsection{Step 3: Client-level Data Generation and Augmentation}
In this step, the clients leverage a state-of-the-art pre-trained SD~\cite{rombach2022high} to synthesize images in a manner that effectively renders their local data IID.
Specifically, for client $k$, let the class labels it owns be $\mathcal{Y}_{in} = \{y | y \in \cD_k\}$, and the class labels it does not own be $\mathcal{Y}_{out} = \{y | y \notin \cD_k\}$. The total class labels are denoted as $\mathcal{Y} = \mathcal{Y}_{in} \cup \mathcal{Y}_{out}$.
The client determines the maximum number of images per class it has across all classes it owns $n_{max}$, i.e.,  $n_{max} = \maxx{y \in \mathcal{Y}_{in}} |\{y \in D^i_{train}\}|$, and generate $n_{max} - \vert\{y \in D_k\}\vert$ synthetic data for the classes it owns and $n_{max}$ synthetic data for the classes it does not own. This ensures that the final augmented local dataset $D_{aug}$, which combines the entire generated synthetic dataset $D_{syn}$ with the local dataset $D_{local}$ (i.e., $D_{aug} = D_{local} \cup D_{syn}$), becomes IID. To illustrate this further, consider an example where a client has 250 images of cats and 100 images of dogs, and the total global classes are cat, dog, and frog. The generated synthetic dataset for this client will include 150 images of dogs and 250 images of frogs. Figure~\ref{fig2} visualizes the local data distribution of clients before and after augmentation for a CIFAR-10 dataset example. After clients augmented their local dataset, they start the federation and locally train the global model on their augmented local datasets $D_{aug}$. Algorithm~\ref{alg:Gen-FedSD} outlines the dataset generation process for each client.

\textbf{Remarks.} It is noteworthy to mention the following remarks:
\begin{itemize}
    \item The privacy implications of our proposed method are equivalent to FedAvg and most existing FL methods that require model exchanges between the server and clients. Notably, sharing the total classes' names with the clients does not pose any privacy issues. In FL, the server possesses the knowledge of the total classes' names in order to design the neural network architecture but lacks any information about the local private dataset of each client. In our method, the server only shares the total classes' names with the clients, and clients construct their local synthesized data based on their specific data distribution disparities.
    \item On-device inference of foundational models such as CLIP and SD is feasible. Recently, Qualcomm Technologies, Inc. has publicly released a demo for on-device inference of Stable Diffusion  \footnote{\url{https://www.qualcomm.com/news/onq/2023/02/worlds-first-on-device-demonstration-of-stable-diffusion-on-android}}.
\end{itemize}

\begin{algorithm}[t]
\caption{Data Generation for Each Client}
\label{alg:Gen-FedSD}

\begin{algorithmic}[1]
\REQUIRE Stable Diffusion Model $\text{SD}$, client $k$, $\mathcal{Y}$, $\mathcal{Y}_{in}$, $\mathcal{Y}_{out}$, prompt templates $\mathcal{M}$.
\STATE $n_{max} \leftarrow \maxx{y \in \mathcal{Y}_{in}} |\{y \in D_k\}|$ \;
\STATE $D_{syn} \leftarrow \emptyset$
\FOR {each $y \in \mathcal{Y}$}
\IF{$y \in \mathcal{Y}_{out}$}
\STATE $D \leftarrow$ generate $n_{max}$ synthetic data by prompting $\text{SD}(M(y))$
\ELSIF{$y \in \mathcal{Y}_{in}$}
\STATE $D \leftarrow$ generate $n_{max} - \vert\{y \in D_k\}\vert$ synthetic data by prompting $\text{SD}(M(y))$
\ENDIF
\STATE $D_{syn} \leftarrow D_{syn} \cup D$
\ENDFOR
\end{algorithmic}
\end{algorithm}
\section{Experiments} \label{sec:experiments}

\subsection{Experimental Setup}

\textbf{Dataset and Architecture.} We conduct experiments on CIFAR-10 and CIFAR-100~\cite{krizhevsky2009learning} datasets. CIFAR-10 contains 50,000 32$\times$32 RGB images across 10 different classes with a 10,000 test dataset. CIFAR-100 contains 50,000 32$\times$32 RGB images of 100 distinct objects with a 10,000 test dataset. We adopt the smallest version of compact convolutional transformer (CCT-2)~\cite{hassani2021escaping} in our experiments for both CIFAR10 and CIFAR-100 datasets.

\textbf{Compared Methods and Evaluation Metric.} We compare our method against four state-of-the-art FL algorithms, including FedAvg~\cite{mcmahan2017communication}, FedProx~\cite{li2020federated}, FedNova~\cite{wang2020tackling}, and Scaffold~\cite{karimireddy2020scaffold}. We use the average global model's top-1 accuracy on the test dataset over the last $10$ rounds as the evaluation metric~\cite{morafah2023practical}. We run each experiment 3 times with different random seeds and report the mean and standard deviation.

\textbf{FL Settings.} For both CIFAR-10 and CIFAR-100 datasets, we simulate a federated learning setting with 100 clients, sample rate 0.1, local batch size 64, and AdamW local optimizer. The number of communication rounds for CIFAR-10 and CIFAR-100 is 100 and 200, respectively. We also fix the number local updates to 150 and 100 for CIFAR-10 and CIFAR-100, respectively~\footnote{Fixing the number of local updates is an algorithmic choice in our setting to have a fair comparison.}. We realize data heterogeneity using Dirichlet distribution with concentration paramter $\alpha$, where lower $\alpha$ results into more heterogeneous data partitions. More details about Dirichlet-based data partitioning can found in~\cite{li2021model, morafah2023practical}.

\textbf{Image Generation Settings.} We download the ``CompVis/stable-diffusion-v1-4'' pre-trained Stable Diffusion model checkpoint to generate the images. We adopt 20 inference steps and a guidance scale of 7.

\textbf{Prompt Templates.} Data diversity has been shown to enhance model generalization in both visual~\cite{chen2020simple} and language domains~\cite{radford2019language}. Employing diverse prompts is a method to generate a wider variety of images~\cite{shipard2023diversity, he2022synthetic}. In our experiment, we utilize two different prompt designs: (1) Without a diverse prompt design, i.e., employing a fixed template ('a photo of a {class}'). (2) With a diverse prompt design, where for each image generation, a prompt template is randomly selected from a pool of options~\footnote{The pool of prompt options is adopted from the original CLIP implementation, available at \url{https://github.com/openai/CLIP/blob/main/data/prompts.md}.}.

\textbf{Implementation.} We implement our code using the FedZoo open source library~\cite{morafah2023practical} in PyTorch. For Stable Diffusion, we use the open source Diffusers~\cite{von-platen-etal-2022-diffusers} library from Hugging Face.

\subsection{Performance Evaluation}

Tables~\ref{tab-cifar10-main} and~\ref{tab-cifar100-main} show the results on CIFAR-10 and CIFAR-100 for the two different prompt templates we discussed, respectively. We can make several interesting observations. 

All vanilla FL baselines suffer significantly from data heterogeneity, and the performance degradation is more significant for higher heterogeneity levels. Furthermore, vanilla-FedProx, -FedNova, and -Scaffold are more severely affected by data heterogeneity compared to FedAvg, specially on CIFAR-100 dataset. This further points out the importance of addressing
data heterogeneity problem in federated learning. Surprisingly, Gen-FedSD by simply augmenting the clients local dataset with synthetic images to make them IID substantially enhances the performance for each FL baseline. For example, the performance of Scaffold has been improved by $34\%$ on CIFAR-100 Dir(0.05), and FedProx has been improved by $41\%$ on CIFAR-10 Dir(0.1). This clearly demonstrate the effectiveness of Gen-FedSD to address the data heterogeneity problem by successfully filling in the clients' unique local distribution mismatches. Another interesting observation is that while vanilla-FedProx, -FedNova, and -Scaffold underperform vanilla-FedAvg, they outperform FedAvg when Gen-FedSD is applied on many cases. Additionally, it is worth highlighting that Ged-FedSD with diverse prompt templates achieves better performance compared to without diverse prompt templates on many cases. This verifies that diverse prompt templates can result into diversified data generation and enhanced generalization of the models.

\newcolumntype{g}{>{\columncolor{mylavender}}l}
\begin{table}[h]
  \caption{Performance results on CIFAR-10 dataset.}
  \label{tab-cifar10-main}
  \centering
  \resizebox{0.99\linewidth}{!}{
  \begin{tabular}{lg|ggg}
    \toprule
    \rowcolor{white} \multicolumn{1}{l}{Baseline} & Vanilla/Gen-FedSD & \multicolumn{1}{c}{Dir (0.5)} & \multicolumn{1}{c}{Dir (0.1)} & \multicolumn{1}{c}{Dir (0.05)} \\
    \midrule
    \rowcolor{white} \multirow{3}{*}{FedAvg} & Vanilla & $58.57\pm0.77$\% & $37.98\pm 1.53$\% & $25.78\pm 0.94$\%\\
      & +Gen-FedSD (w/o diversity) & $\bm{70.70 \pm 0.62}$\% & $\bm{58.69 \pm 1.71}$\% & $\bm{54.38 \pm1.93}$\%\\
      & +Gen-FedSD (w diversity) & $\bm{71.41 \pm 0.45}$\% & $\bm{61.21 \pm 1.74}$\% & $\bm{56.42 \pm 0.35}$\%\\
      \midrule
    \rowcolor{white} \multirow{3}{*}{FedProx} & Vanilla & $27.42 \pm 1.84$\% & $20.51 \pm 1.79$\% & $18.05 \pm 1.98$\%\\
      & +Gen-FedSD (w/o diversity) & $\bm{70.26 \pm 0.14}$\% & $\bm{61.59 \pm 0.98}$\% & $\bm{54.85 \pm 0.95}$\%\\
      & +Gen-FedSD (w diversity) & $\bm{71.01 \pm 0.17}$\% & $\bm{61.43 \pm 1.27}$\% & $\bm{59.41 \pm 0.76}$\%\\
      \midrule
      \rowcolor{white} \multirow{3}{*}{FedNova} & Vanilla & $28.52 \pm 1.01$\% & $24.61 \pm 1.24$\% & $19.37 \pm 2.12$\%\\
      & +Gen-FedSD (w/o diversity) & $\bm{70.06 \pm 0.34}$\% & $\bm{61.45 \pm 1.29}$\% & $\bm{55.85 \pm 2.19}$\%\\
      & +Gen-FedSD (w diversity) & $\bm{71.06 \pm 0.10}$\% & $\bm{63.82 \pm 0.09}$\% & $\bm{57.92 \pm 0.97}$\%\\
      \midrule
    \rowcolor{white} \multirow{3}{*}{Scaffold} & Vanilla & $26.65 \pm 0.45$\% & $17.71 \pm 1.71$\% & $14.14 \pm 1.48$\% \\
      & +Gen-FedSD (w/o diversity) & $\bm{76.39 \pm 0.10}$\% & $\bm{67.46 \pm 1.17}$\% & $\bm{59.04 \pm 2.35}$\% \\
      & +Gen-FedSD (w diversity) & $\bm{76.61 \pm 0.26}$\% & $\bm{67.09 \pm 0.04}$\% & $\bm{61.91 \pm 3.31}$\% \\
    \bottomrule
  \end{tabular}
  }
\end{table}

\newcolumntype{g}{>{\columncolor{mylavender}}l}
\begin{table}[h]
  \caption{Performance results on CIFAR-100 dataset.}
  \label{tab-cifar100-main}
  \centering
  \resizebox{0.99\linewidth}{!}{
  \begin{tabular}{lg|ggg}
    \toprule
    \rowcolor{white} \multicolumn{1}{l}{Baseline} & Vanilla/Gen-FedSD & \multicolumn{1}{c}{Dir (0.5)} & \multicolumn{1}{c}{Dir (0.1)} & \multicolumn{1}{c}{Dir (0.05)} \\
    \midrule
    \rowcolor{white} \multirow{3}{*}{FedAvg} & Vanilla & $40.36 \pm 0.39$\% & $30.38 \pm 0.78$\% & $24.52 \pm 0.37$\% \\
      & +Gen-FedSD (w/o diversity) & $\bm{46.45\pm0.30}$\% & $\bm{36.23 \pm 0.78 \%}$ & $\bm{31.42\pm 0.24}$\%\\
      & +Gen-FedSD (w diversity) & $\bm{46.91 \pm 0.50}$\% & $\bm{37.06 \pm 0.20}$\% & $\bm{31.91 \pm 0.25}$\%\\
      \midrule
    \rowcolor{white} \multirow{3}{*}{FedProx} & Vanilla & $10.39 \pm 1.27$\% & $8.12 \pm 0.72$\% & $6.57 \pm 0.07$\%\\
      & +Gen-FedSD (w/o diversity) & $\bm{47.36 \pm 0.03}$\% & $\bm{39.69 \pm 0.18}$\% & $\bm{35.29 \pm 0.80}$\%\\
        & +Gen-FedSD (w diversity) & $\bm{47.44 \pm 0.10}$\% & $\bm{39.22 \pm 0.31}$\% & $\bm{36.14 \pm 0.55}$\%\\
        \midrule
      \rowcolor{white} \multirow{3}{*}{FedNova} & Vanilla & $8.71 \pm 1.31$\% & $7.48 \pm 0.48$\% & $6.83 \pm 0.19$\%\\
      & +Gen-FedSD (w/o diversity) & $\bm{46.98 \pm 0.03}$\% & $\bm{39.24 \pm 0.41}$\% & $\bm{35.86 \pm 0.14}$\%\\
      & +Gen-FedSD (w diversity) & $\bm{47.44 \pm 0.50}$\% & $\bm{39.83 \pm 0.33}$\% & $\bm{36.10 \pm 0.27}$\%\\
      \midrule
    \rowcolor{white} \multirow{3}{*}{Scaffold} & Vanilla & $4.49 \pm 0.33$\% & $3.32 \pm 0.05$\% & $2.63 \pm 0.22$\% \\
      & +Gen-FedSD (w/o diversity) & $\bm{52.25 \pm 0.33}$\% & $\bm{42.70 \pm 0.61}$\% & $\bm{36.79 \pm 0.11}$\% \\
      & +Gen-FedSD (w diversity) & $\bm{52.43 \pm 0.06}$\% & $\bm{43.77 \pm 0.02}$\% & $\bm{38.42 \pm 0.02}$\% \\
    \bottomrule
  \end{tabular}
  }
\end{table}

\newcolumntype{g}{>{\columncolor{mylavender}}l}
\begin{table}[hb]
  \caption{Communication costs required to reach the designated target accuracy for CIFAR-10 dataset.}
  \label{tab-cifar10-comm}
  \centering
  \resizebox{0.99\linewidth}{!}{
  \begin{tabular}{lg|ggg}
    \toprule
    \rowcolor{white} \multicolumn{1}{l}{Baseline} & Vanilla/Gen-FedSD & \multicolumn{1}{c}{Dir (0.5)} & \multicolumn{1}{c}{Dir (0.1)} & \multicolumn{1}{c}{Dir (0.05)} \\
    \midrule
    \rowcolor{white} \multirow{3}{*}{FedAvg} & Vanilla & $567.44$ MB & $408.56$ MB & $340.46$ MB \\
      & +Gen-FedSD (w/o diversity) & $\bm{22.69}$ MB & $\bm{22.69}$ MB & $\bm{22.69}$ MB\\
        & +Gen-FedSD (w diversity) & $\bm{22.69}$ MB & $\bm{22.69}$ MB & $\bm{22.69}$ MB \\
    \midrule
    \rowcolor{white} \multirow{3}{*}{FedProx} & Vanilla & $1588.84$ MB & $1134.89$ MB & $499.35$ MB \\
      & +Gen-FedSD (w/o diversity) & $\bm{22.69}$ MB & $\bm{22.69}$ MB & $\bm{22.69}$ MB\\
      & +Gen-FedSD (w diversity) & $\bm{22.69}$ MB & $\bm{22.69}$ MB & $\bm{22.69}$ MB\\
      \midrule
      \rowcolor{white} \multirow{3}{*}{FedNova} & Vanilla & $907.91$ MB & $453.95$ MB & $794.42$ MB \\
      & +Gen-FedSD (w/o diversity) & $\bm{22.69}$ MB & $\bm{22.69}$ MB  & $\bm{22.69}$ MB\\
      & +Gen-FedSD (w diversity) & $\bm{22.69}$ MB & $\bm{22.69}$ MB & $\bm{22.69}$ MB\\
      \midrule
    \rowcolor{white} \multirow{3}{*}{Scaffold} & Vanilla & $794.42$ MB & $1361.87$ MB & $1134.89$ MB \\
      & +Gen-FedSD (w/o diversity) & $\bm{22.69}$ MB & $\bm{22.69}$ MB  & $\bm{22.69}$ MB\\
      & +Gen-FedSD (w diversity) & $\bm{22.69}$ MB & $\bm{22.69}$ MB & $\bm{22.69}$ MB\\
    \bottomrule
  \end{tabular}
  }
\end{table}

\subsection{Communication Cost}
To measure the communication cost, we set specific target accuracies and evaluate each baseline on the number of communication costs required to reach these targets. Specifically, we set the target accuracies to 25\% for Dir(0.5), 20\% for Dir(0.1), and 18\% for Dir(0.05). Table~\ref{tab-cifar10-comm} presents the communication costs of the baselines on the CIFAR-10 dataset in MB~\footnote{The communication cost evaluation for CIFAR-100 dataset cannot be accomplished since the performance results of all the baselines except FedAvg is very poor. Hence, we only conduct the evaluation on the CIFAR-10 dataset.} As evident from the table, Gen-FedSD demonstrates significant reductions in communication costs across all levels of data heterogeneity compared to the vanilla baselines.
\section{Conclusion and Future Works} \label{sec:conclusion}
In this work, we introduce ``Gen-FedSD,'' a novel data augmentation technique leveraging pre-trained state-of-the-art Stable Diffusion to tackle the challenge of data heterogeneity in federated learning. Gen-FedSD customizes images for each client based on its unique data distribution disparities, aiming to make it independent and identically distributed (IID) with respect to the global data distribution. Our experiments demonstrate that Gen-FedSD substantially enhances the performance of several state-of-the-art federated learning algorithms across diverse datasets and Non-IID scenarios.

For future work, it would be valuable to explore different prompt engineering techniques and factors affecting image generation quality. Further investigations into other sources of non-iidness and how the method could be adapted would be intriguing extensions for future research.

\appendix
\section{Supplementary Materials}
\subsection{Further Experimental Settings}

\textbf{FL Settings and Hyper-parameters.} Tables~\ref{hyper-parameters-cifar10} and~\ref{hyper-parameters-cifar100} summarize the details of FL settings and hyper-parameters that we used in our federated learning experiments for CIFAR-10 and CIFAR-100, respectively.

\begin{table}[h]
    \centering
    \caption{Federated learning settings and hyper-parameters for CIFAR-10.}
    \label{hyper-parameters-cifar10}
    \resizebox{0.30\textwidth}{!}{
    \begin{tabular}{c|c}
        \toprule
         Name & Value\\
         \midrule
         Dataset & CIFAR-10\\ 
         Architecture & CCT-2\\
         \# parameters & 283,723\\
         Number of Clients & 100\\
         Sample rate & 0.1\\
         local batch size & 64\\
         local update iterations & 150\\
         local optimizer & AdamW\\
         local learning rate & 5e-4\\
         local weight decay & 3e-2\\
         \bottomrule
    \end{tabular}
    }
\end{table}

\begin{table}[h]
    \centering
    \caption{Federated learning settings and hyper-parameters for CIFAR-100.}
    \label{hyper-parameters-cifar100}
    \resizebox{0.30\textwidth}{!}{
    \begin{tabular}{c|c}
        \toprule
         Name & Value\\
         \midrule
         Dataset & CIFAR-100\\ 
         Architecture & CCT-2\\
         \# parameters & 295,333\\
         rounds & 200 \\
         Number of Clients & 100\\
         Sample rate & 0.1\\
         local batch size & 64\\
         local update iterations & 100\\
         local optimizer & AdamW\\
         local learning rate & 6e-4\\
         local weight decay & 6e-2\\
         \bottomrule
    \end{tabular}
    }
\end{table}

\textbf{Dataset Generation Setting.} We utilized Stable Diffusion~\cite{rombach2022high} model and the pre-trained stable-diffusion-v1-4 checkpoint from huggingface available at ~\url{https://huggingface.co/CompVis/stable-diffusion} to generate data. We used the inference code in the diffusers~\cite{von2022diffusers} library. Table~\ref{hyper-parameters-stable-diffusion} summarizes the hyper-parameters we used to generated data using Stable Diffusion. 

\begin{table}[h]
    \centering
    \caption{Inference settings for data generationg.}
    \label{hyper-parameters-stable-diffusion}
    \resizebox{0.40\textwidth}{!}{
    \begin{tabular}{c|c}
        \toprule
         Name & Value\\
         \midrule
         Model & Stable Diffusion\\ 
         Pre-trained & ``CompVis/stable-diffusion-v1-4''\\
         torch\_dtype & torch.float16\\
         Scheduler & DPMSolverMultistepScheduler \\
         num\_inference\_stepss & 20\\
         guidance\_scale & 7\\
         \bottomrule
    \end{tabular}
    }
\end{table}

\textbf{Prompt Templates.} We followed original CLIP prompt templates~\footnote{The prompts are available at the official Github repository:~\url{https://github.com/openai/CLIP/blob/main/data/prompts.md}.}. Table~\ref{cifar-10-prompts} summarizes the prompt templates that we used for CIFAR-10 and CIFAR-100 datasets. 

\begin{table}[htbp]
\caption{CIFAR-10 and CIFAR-100 Prompt Templates}\label{cifar-10-prompts}
\begin{tcolorbox}[colback=white!5!white,colframe=black!75!black]
    'a photo of a \{\texttt{class}\}.',
    'a blurry photo of a \{\texttt{class}\}.',
    'a black and white photo of a \{\texttt{class}\}.',
    'a low contrast photo of a \{\texttt{class}\}.',
    'a high contrast photo of a \{\texttt{class}\}.',
    'a bad photo of a \{\texttt{class}\}.',
    'a good photo of a \{\texttt{class}\}.',
    'a photo of a small \{\texttt{class}\}.',
    'a photo of a big \{\texttt{class}\}.',
    'a photo of the \{\texttt{class}\}.',
    'a blurry photo of the \{\texttt{class}\}.',
    'a black and white photo of the \{\texttt{class}\}.',
    'a low contrast photo of the \{\texttt{class}\}.',
    'a high contrast photo of the \{\texttt{class}\}.',
    'a bad photo of the \{\texttt{class}\}.',
    'a good photo of the \{\texttt{class}\}.',
    'a photo of the small \{\texttt{class}\}.',
    'a photo of the big \{\texttt{class}\}.',
\end{tcolorbox}
\end{table}

\newpage
\bibliographystyle{ACM-Reference-Format}
\bibliography{refs}


\begin{thebibliography}{68}


\ifx \showCODEN    \undefined \def \showCODEN     #1{\unskip}     \fi
\ifx \showDOI      \undefined \def \showDOI       #1{#1}\fi
\ifx \showISBNx    \undefined \def \showISBNx     #1{\unskip}     \fi
\ifx \showISBNxiii \undefined \def \showISBNxiii  #1{\unskip}     \fi
\ifx \showISSN     \undefined \def \showISSN      #1{\unskip}     \fi
\ifx \showLCCN     \undefined \def \showLCCN      #1{\unskip}     \fi
\ifx \shownote     \undefined \def \shownote      #1{#1}          \fi
\ifx \showarticletitle \undefined \def \showarticletitle #1{#1}   \fi
\ifx \showURL      \undefined \def \showURL       {\relax}        \fi
\providecommand\bibfield[2]{#2}
\providecommand\bibinfo[2]{#2}
\providecommand\natexlab[1]{#1}
\providecommand\showeprint[2][]{arXiv:#2}

\bibitem[Azizi et~al\mbox{.}(2023)]%
        {azizi2023synthetic}
\bibfield{author}{\bibinfo{person}{Shekoofeh Azizi}, \bibinfo{person}{Simon Kornblith}, \bibinfo{person}{Chitwan Saharia}, \bibinfo{person}{Mohammad Norouzi}, {and} \bibinfo{person}{David~J Fleet}.} \bibinfo{year}{2023}\natexlab{}.
\newblock \showarticletitle{Synthetic data from diffusion models improves imagenet classification}.
\newblock \bibinfo{journal}{\emph{arXiv preprint arXiv:2304.08466}} (\bibinfo{year}{2023}).
\newblock


\bibitem[Bansal and Grover(2023)]%
        {bansal2023leaving}
\bibfield{author}{\bibinfo{person}{Hritik Bansal} {and} \bibinfo{person}{Aditya Grover}.} \bibinfo{year}{2023}\natexlab{}.
\newblock \showarticletitle{Leaving reality to imagination: Robust classification via generated datasets}.
\newblock \bibinfo{journal}{\emph{arXiv preprint arXiv:2302.02503}} (\bibinfo{year}{2023}).
\newblock


\bibitem[Baranchuk et~al\mbox{.}(2021)]%
        {baranchuk2021label}
\bibfield{author}{\bibinfo{person}{Dmitry Baranchuk}, \bibinfo{person}{Ivan Rubachev}, \bibinfo{person}{Andrey Voynov}, \bibinfo{person}{Valentin Khrulkov}, {and} \bibinfo{person}{Artem Babenko}.} \bibinfo{year}{2021}\natexlab{}.
\newblock \showarticletitle{Label-efficient semantic segmentation with diffusion models}.
\newblock \bibinfo{journal}{\emph{arXiv preprint arXiv:2112.03126}} (\bibinfo{year}{2021}).
\newblock


\bibitem[Cao et~al\mbox{.}(2024)]%
        {cao2024survey}
\bibfield{author}{\bibinfo{person}{Hanqun Cao}, \bibinfo{person}{Cheng Tan}, \bibinfo{person}{Zhangyang Gao}, \bibinfo{person}{Yilun Xu}, \bibinfo{person}{Guangyong Chen}, \bibinfo{person}{Pheng-Ann Heng}, {and} \bibinfo{person}{Stan~Z Li}.} \bibinfo{year}{2024}\natexlab{}.
\newblock \showarticletitle{A Survey on Generative Diffusion Models}.
\newblock \bibinfo{journal}{\emph{IEEE Transactions on Knowledge and Data Engineering}} (\bibinfo{year}{2024}).
\newblock


\bibitem[Chen et~al\mbox{.}(2019)]%
        {chen2019federated}
\bibfield{author}{\bibinfo{person}{Mingqing Chen}, \bibinfo{person}{Rajiv Mathews}, \bibinfo{person}{Tom Ouyang}, {and} \bibinfo{person}{Fran{\c{c}}oise Beaufays}.} \bibinfo{year}{2019}\natexlab{}.
\newblock \showarticletitle{Federated learning of out-of-vocabulary words}.
\newblock \bibinfo{journal}{\emph{arXiv preprint arXiv:1903.10635}} (\bibinfo{year}{2019}).
\newblock


\bibitem[Chen et~al\mbox{.}(2020)]%
        {chen2020simple}
\bibfield{author}{\bibinfo{person}{Ting Chen}, \bibinfo{person}{Simon Kornblith}, \bibinfo{person}{Mohammad Norouzi}, {and} \bibinfo{person}{Geoffrey Hinton}.} \bibinfo{year}{2020}\natexlab{}.
\newblock \showarticletitle{A simple framework for contrastive learning of visual representations}. In \bibinfo{booktitle}{\emph{International conference on machine learning}}. PMLR, \bibinfo{pages}{1597--1607}.
\newblock


\bibitem[Collins et~al\mbox{.}(2021)]%
        {collins2021exploiting}
\bibfield{author}{\bibinfo{person}{Liam Collins}, \bibinfo{person}{Hamed Hassani}, \bibinfo{person}{Aryan Mokhtari}, {and} \bibinfo{person}{Sanjay Shakkottai}.} \bibinfo{year}{2021}\natexlab{}.
\newblock \showarticletitle{Exploiting shared representations for personalized federated learning}. In \bibinfo{booktitle}{\emph{International conference on machine learning}}. PMLR, \bibinfo{pages}{2089--2099}.
\newblock


\bibitem[Deng et~al\mbox{.}(2020)]%
        {deng2020adaptive}
\bibfield{author}{\bibinfo{person}{Yuyang Deng}, \bibinfo{person}{Mohammad~Mahdi Kamani}, {and} \bibinfo{person}{Mehrdad Mahdavi}.} \bibinfo{year}{2020}\natexlab{}.
\newblock \showarticletitle{Adaptive personalized federated learning}.
\newblock \bibinfo{journal}{\emph{arXiv preprint arXiv:2003.13461}} (\bibinfo{year}{2020}).
\newblock


\bibitem[Dhariwal and Nichol(2021)]%
        {dhariwal2021diffusion}
\bibfield{author}{\bibinfo{person}{Prafulla Dhariwal} {and} \bibinfo{person}{Alexander Nichol}.} \bibinfo{year}{2021}\natexlab{}.
\newblock \showarticletitle{Diffusion models beat gans on image synthesis}.
\newblock \bibinfo{journal}{\emph{Advances in neural information processing systems}}  \bibinfo{volume}{34} (\bibinfo{year}{2021}), \bibinfo{pages}{8780--8794}.
\newblock


\bibitem[Feng et~al\mbox{.}(2020)]%
        {feng2020pmf}
\bibfield{author}{\bibinfo{person}{Jie Feng}, \bibinfo{person}{Can Rong}, \bibinfo{person}{Funing Sun}, \bibinfo{person}{Diansheng Guo}, {and} \bibinfo{person}{Yong Li}.} \bibinfo{year}{2020}\natexlab{}.
\newblock \showarticletitle{PMF: A privacy-preserving human mobility prediction framework via federated learning}.
\newblock \bibinfo{journal}{\emph{Proceedings of the ACM on Interactive, Mobile, Wearable and Ubiquitous Technologies}} \bibinfo{volume}{4}, \bibinfo{number}{1} (\bibinfo{year}{2020}), \bibinfo{pages}{1--21}.
\newblock


\bibitem[Goetz and Tewari(2020)]%
        {goetz2020federated}
\bibfield{author}{\bibinfo{person}{Jack Goetz} {and} \bibinfo{person}{Ambuj Tewari}.} \bibinfo{year}{2020}\natexlab{}.
\newblock \showarticletitle{Federated learning via synthetic data}.
\newblock \bibinfo{journal}{\emph{arXiv preprint arXiv:2008.04489}} (\bibinfo{year}{2020}).
\newblock


\bibitem[Gu et~al\mbox{.}(2023)]%
        {gu2023systematic}
\bibfield{author}{\bibinfo{person}{Jindong Gu}, \bibinfo{person}{Zhen Han}, \bibinfo{person}{Shuo Chen}, \bibinfo{person}{Ahmad Beirami}, \bibinfo{person}{Bailan He}, \bibinfo{person}{Gengyuan Zhang}, \bibinfo{person}{Ruotong Liao}, \bibinfo{person}{Yao Qin}, \bibinfo{person}{Volker Tresp}, {and} \bibinfo{person}{Philip Torr}.} \bibinfo{year}{2023}\natexlab{}.
\newblock \showarticletitle{A systematic survey of prompt engineering on vision-language foundation models}.
\newblock \bibinfo{journal}{\emph{arXiv preprint arXiv:2307.12980}} (\bibinfo{year}{2023}).
\newblock


\bibitem[Haddadpour and Mahdavi(2019)]%
        {haddadpour2019convergence}
\bibfield{author}{\bibinfo{person}{Farzin Haddadpour} {and} \bibinfo{person}{Mehrdad Mahdavi}.} \bibinfo{year}{2019}\natexlab{}.
\newblock \showarticletitle{On the convergence of local descent methods in federated learning}.
\newblock \bibinfo{journal}{\emph{arXiv preprint arXiv:1910.14425}} (\bibinfo{year}{2019}).
\newblock


\bibitem[Hanzely et~al\mbox{.}(2020)]%
        {hanzely2020lower}
\bibfield{author}{\bibinfo{person}{Filip Hanzely}, \bibinfo{person}{Slavom{\'\i}r Hanzely}, \bibinfo{person}{Samuel Horv{\'a}th}, {and} \bibinfo{person}{Peter Richt{\'a}rik}.} \bibinfo{year}{2020}\natexlab{}.
\newblock \showarticletitle{Lower bounds and optimal algorithms for personalized federated learning}.
\newblock \bibinfo{journal}{\emph{Advances in Neural Information Processing Systems}}  \bibinfo{volume}{33} (\bibinfo{year}{2020}), \bibinfo{pages}{2304--2315}.
\newblock


\bibitem[Hassani et~al\mbox{.}(2021)]%
        {hassani2021escaping}
\bibfield{author}{\bibinfo{person}{Ali Hassani}, \bibinfo{person}{Steven Walton}, \bibinfo{person}{Nikhil Shah}, \bibinfo{person}{Abulikemu Abuduweili}, \bibinfo{person}{Jiachen Li}, {and} \bibinfo{person}{Humphrey Shi}.} \bibinfo{year}{2021}\natexlab{}.
\newblock \showarticletitle{Escaping the big data paradigm with compact transformers}.
\newblock \bibinfo{journal}{\emph{arXiv preprint arXiv:2104.05704}} (\bibinfo{year}{2021}).
\newblock


\bibitem[He et~al\mbox{.}(2022)]%
        {he2022synthetic}
\bibfield{author}{\bibinfo{person}{Ruifei He}, \bibinfo{person}{Shuyang Sun}, \bibinfo{person}{Xin Yu}, \bibinfo{person}{Chuhui Xue}, \bibinfo{person}{Wenqing Zhang}, \bibinfo{person}{Philip Torr}, \bibinfo{person}{Song Bai}, {and} \bibinfo{person}{Xiaojuan Qi}.} \bibinfo{year}{2022}\natexlab{}.
\newblock \showarticletitle{Is synthetic data from generative models ready for image recognition?}
\newblock \bibinfo{journal}{\emph{arXiv preprint arXiv:2210.07574}} (\bibinfo{year}{2022}).
\newblock


\bibitem[Ho et~al\mbox{.}(2020)]%
        {ho2020denoising}
\bibfield{author}{\bibinfo{person}{Jonathan Ho}, \bibinfo{person}{Ajay Jain}, {and} \bibinfo{person}{Pieter Abbeel}.} \bibinfo{year}{2020}\natexlab{}.
\newblock \showarticletitle{Denoising diffusion probabilistic models}.
\newblock \bibinfo{journal}{\emph{Advances in neural information processing systems}}  \bibinfo{volume}{33} (\bibinfo{year}{2020}), \bibinfo{pages}{6840--6851}.
\newblock


\bibitem[Ho et~al\mbox{.}(2022)]%
        {ho2022cascaded}
\bibfield{author}{\bibinfo{person}{Jonathan Ho}, \bibinfo{person}{Chitwan Saharia}, \bibinfo{person}{William Chan}, \bibinfo{person}{David~J Fleet}, \bibinfo{person}{Mohammad Norouzi}, {and} \bibinfo{person}{Tim Salimans}.} \bibinfo{year}{2022}\natexlab{}.
\newblock \showarticletitle{Cascaded diffusion models for high fidelity image generation}.
\newblock \bibinfo{journal}{\emph{The Journal of Machine Learning Research}} \bibinfo{volume}{23}, \bibinfo{number}{1} (\bibinfo{year}{2022}), \bibinfo{pages}{2249--2281}.
\newblock


\bibitem[Hu et~al\mbox{.}(2022)]%
        {hu2022fedsynth}
\bibfield{author}{\bibinfo{person}{Shengyuan Hu}, \bibinfo{person}{Jack Goetz}, \bibinfo{person}{Kshitiz Malik}, \bibinfo{person}{Hongyuan Zhan}, \bibinfo{person}{Zhe Liu}, {and} \bibinfo{person}{Yue Liu}.} \bibinfo{year}{2022}\natexlab{}.
\newblock \showarticletitle{Fedsynth: Gradient compression via synthetic data in federated learning}.
\newblock \bibinfo{journal}{\emph{arXiv preprint arXiv:2204.01273}} (\bibinfo{year}{2022}).
\newblock


\bibitem[Jeong et~al\mbox{.}(2018)]%
        {jeong2018communication}
\bibfield{author}{\bibinfo{person}{Eunjeong Jeong}, \bibinfo{person}{Seungeun Oh}, \bibinfo{person}{Hyesung Kim}, \bibinfo{person}{Jihong Park}, \bibinfo{person}{Mehdi Bennis}, {and} \bibinfo{person}{Seong-Lyun Kim}.} \bibinfo{year}{2018}\natexlab{}.
\newblock \showarticletitle{Communication-efficient on-device machine learning: Federated distillation and augmentation under non-iid private data}.
\newblock \bibinfo{journal}{\emph{arXiv preprint arXiv:1811.11479}} (\bibinfo{year}{2018}).
\newblock


\bibitem[Kairouz et~al\mbox{.}(2021)]%
        {kairouz2021advances}
\bibfield{author}{\bibinfo{person}{Peter Kairouz}, \bibinfo{person}{H~Brendan McMahan}, \bibinfo{person}{Brendan Avent}, \bibinfo{person}{Aur{\'e}lien Bellet}, \bibinfo{person}{Mehdi Bennis}, \bibinfo{person}{Arjun~Nitin Bhagoji}, \bibinfo{person}{Kallista Bonawitz}, \bibinfo{person}{Zachary Charles}, \bibinfo{person}{Graham Cormode}, \bibinfo{person}{Rachel Cummings}, {et~al\mbox{.}}} \bibinfo{year}{2021}\natexlab{}.
\newblock \showarticletitle{Advances and open problems in federated learning}.
\newblock \bibinfo{journal}{\emph{Foundations and Trends{\textregistered} in Machine Learning}} \bibinfo{volume}{14}, \bibinfo{number}{1--2} (\bibinfo{year}{2021}), \bibinfo{pages}{1--210}.
\newblock


\bibitem[Karimireddy et~al\mbox{.}(2020)]%
        {karimireddy2020scaffold}
\bibfield{author}{\bibinfo{person}{Sai~Praneeth Karimireddy}, \bibinfo{person}{Satyen Kale}, \bibinfo{person}{Mehryar Mohri}, \bibinfo{person}{Sashank Reddi}, \bibinfo{person}{Sebastian Stich}, {and} \bibinfo{person}{Ananda~Theertha Suresh}.} \bibinfo{year}{2020}\natexlab{}.
\newblock \showarticletitle{Scaffold: Stochastic controlled averaging for federated learning}. In \bibinfo{booktitle}{\emph{International conference on machine learning}}. PMLR, \bibinfo{pages}{5132--5143}.
\newblock


\bibitem[Kingma and Welling(2013)]%
        {kingma2013auto}
\bibfield{author}{\bibinfo{person}{Diederik~P Kingma} {and} \bibinfo{person}{Max Welling}.} \bibinfo{year}{2013}\natexlab{}.
\newblock \showarticletitle{Auto-encoding variational bayes}.
\newblock \bibinfo{journal}{\emph{arXiv preprint arXiv:1312.6114}} (\bibinfo{year}{2013}).
\newblock


\bibitem[Krizhevsky et~al\mbox{.}(2009)]%
        {krizhevsky2009learning}
\bibfield{author}{\bibinfo{person}{Alex Krizhevsky}, \bibinfo{person}{Geoffrey Hinton}, {et~al\mbox{.}}} \bibinfo{year}{2009}\natexlab{}.
\newblock \showarticletitle{Learning multiple layers of features from tiny images}.
\newblock  (\bibinfo{year}{2009}).
\newblock


\bibitem[Kulkarni et~al\mbox{.}(2020)]%
        {kulkarni2020survey}
\bibfield{author}{\bibinfo{person}{Viraj Kulkarni}, \bibinfo{person}{Milind Kulkarni}, {and} \bibinfo{person}{Aniruddha Pant}.} \bibinfo{year}{2020}\natexlab{}.
\newblock \showarticletitle{Survey of personalization techniques for federated learning}. In \bibinfo{booktitle}{\emph{2020 Fourth World Conference on Smart Trends in Systems, Security and Sustainability (WorldS4)}}. IEEE, \bibinfo{pages}{794--797}.
\newblock


\bibitem[Li et~al\mbox{.}(2022b)]%
        {li2022srdiff}
\bibfield{author}{\bibinfo{person}{Haoying Li}, \bibinfo{person}{Yifan Yang}, \bibinfo{person}{Meng Chang}, \bibinfo{person}{Shiqi Chen}, \bibinfo{person}{Huajun Feng}, \bibinfo{person}{Zhihai Xu}, \bibinfo{person}{Qi Li}, {and} \bibinfo{person}{Yueting Chen}.} \bibinfo{year}{2022}\natexlab{b}.
\newblock \showarticletitle{Srdiff: Single image super-resolution with diffusion probabilistic models}.
\newblock \bibinfo{journal}{\emph{Neurocomputing}}  \bibinfo{volume}{479} (\bibinfo{year}{2022}), \bibinfo{pages}{47--59}.
\newblock


\bibitem[Li et~al\mbox{.}(2022a)]%
        {li2022federated}
\bibfield{author}{\bibinfo{person}{Qinbin Li}, \bibinfo{person}{Yiqun Diao}, \bibinfo{person}{Quan Chen}, {and} \bibinfo{person}{Bingsheng He}.} \bibinfo{year}{2022}\natexlab{a}.
\newblock \showarticletitle{Federated learning on non-iid data silos: An experimental study}. In \bibinfo{booktitle}{\emph{2022 IEEE 38th International Conference on Data Engineering (ICDE)}}. IEEE, \bibinfo{pages}{965--978}.
\newblock


\bibitem[Li et~al\mbox{.}(2021)]%
        {li2021model}
\bibfield{author}{\bibinfo{person}{Qinbin Li}, \bibinfo{person}{Bingsheng He}, {and} \bibinfo{person}{Dawn Song}.} \bibinfo{year}{2021}\natexlab{}.
\newblock \showarticletitle{Model-contrastive federated learning}. In \bibinfo{booktitle}{\emph{Proceedings of the IEEE/CVF conference on computer vision and pattern recognition}}. \bibinfo{pages}{10713--10722}.
\newblock


\bibitem[Li et~al\mbox{.}(2020)]%
        {li2020federated}
\bibfield{author}{\bibinfo{person}{Tian Li}, \bibinfo{person}{Anit~Kumar Sahu}, \bibinfo{person}{Manzil Zaheer}, \bibinfo{person}{Maziar Sanjabi}, \bibinfo{person}{Ameet Talwalkar}, {and} \bibinfo{person}{Virginia Smith}.} \bibinfo{year}{2020}\natexlab{}.
\newblock \showarticletitle{Federated optimization in heterogeneous networks}.
\newblock \bibinfo{journal}{\emph{Proceedings of Machine learning and systems}}  \bibinfo{volume}{2} (\bibinfo{year}{2020}), \bibinfo{pages}{429--450}.
\newblock


\bibitem[Lin et~al\mbox{.}(2020)]%
        {lin2020ensemble}
\bibfield{author}{\bibinfo{person}{Tao Lin}, \bibinfo{person}{Lingjing Kong}, \bibinfo{person}{Sebastian~U Stich}, {and} \bibinfo{person}{Martin Jaggi}.} \bibinfo{year}{2020}\natexlab{}.
\newblock \showarticletitle{Ensemble distillation for robust model fusion in federated learning}.
\newblock \bibinfo{journal}{\emph{Advances in Neural Information Processing Systems}}  \bibinfo{volume}{33} (\bibinfo{year}{2020}), \bibinfo{pages}{2351--2363}.
\newblock


\bibitem[McMahan et~al\mbox{.}(2017)]%
        {mcmahan2017communication}
\bibfield{author}{\bibinfo{person}{Brendan McMahan}, \bibinfo{person}{Eider Moore}, \bibinfo{person}{Daniel Ramage}, \bibinfo{person}{Seth Hampson}, {and} \bibinfo{person}{Blaise~Aguera y Arcas}.} \bibinfo{year}{2017}\natexlab{}.
\newblock \showarticletitle{Communication-efficient learning of deep networks from decentralized data}. In \bibinfo{booktitle}{\emph{Artificial intelligence and statistics}}. PMLR, \bibinfo{pages}{1273--1282}.
\newblock


\bibitem[Morafah et~al\mbox{.}(2023)]%
        {morafah2023practical}
\bibfield{author}{\bibinfo{person}{Mahdi Morafah}, \bibinfo{person}{Weijia Wang}, {and} \bibinfo{person}{Bill Lin}.} \bibinfo{year}{2023}\natexlab{}.
\newblock \showarticletitle{A Practical Recipe for Federated Learning Under Statistical Heterogeneity Experimental Design}.
\newblock \bibinfo{journal}{\emph{IEEE Transactions on Artificial Intelligence}} (\bibinfo{year}{2023}).
\newblock


\bibitem[Nichol et~al\mbox{.}(2021)]%
        {nichol2021glide}
\bibfield{author}{\bibinfo{person}{Alex Nichol}, \bibinfo{person}{Prafulla Dhariwal}, \bibinfo{person}{Aditya Ramesh}, \bibinfo{person}{Pranav Shyam}, \bibinfo{person}{Pamela Mishkin}, \bibinfo{person}{Bob McGrew}, \bibinfo{person}{Ilya Sutskever}, {and} \bibinfo{person}{Mark Chen}.} \bibinfo{year}{2021}\natexlab{}.
\newblock \showarticletitle{Glide: Towards photorealistic image generation and editing with text-guided diffusion models}.
\newblock \bibinfo{journal}{\emph{arXiv preprint arXiv:2112.10741}} (\bibinfo{year}{2021}).
\newblock


\bibitem[Nichol and Dhariwal(2021)]%
        {nichol2021improved}
\bibfield{author}{\bibinfo{person}{Alexander~Quinn Nichol} {and} \bibinfo{person}{Prafulla Dhariwal}.} \bibinfo{year}{2021}\natexlab{}.
\newblock \showarticletitle{Improved denoising diffusion probabilistic models}. In \bibinfo{booktitle}{\emph{International Conference on Machine Learning}}. PMLR, \bibinfo{pages}{8162--8171}.
\newblock


\bibitem[Radford et~al\mbox{.}(2021)]%
        {radford2021learning}
\bibfield{author}{\bibinfo{person}{Alec Radford}, \bibinfo{person}{Jong~Wook Kim}, \bibinfo{person}{Chris Hallacy}, \bibinfo{person}{Aditya Ramesh}, \bibinfo{person}{Gabriel Goh}, \bibinfo{person}{Sandhini Agarwal}, \bibinfo{person}{Girish Sastry}, \bibinfo{person}{Amanda Askell}, \bibinfo{person}{Pamela Mishkin}, \bibinfo{person}{Jack Clark}, {et~al\mbox{.}}} \bibinfo{year}{2021}\natexlab{}.
\newblock \showarticletitle{Learning transferable visual models from natural language supervision}. In \bibinfo{booktitle}{\emph{International conference on machine learning}}. PMLR, \bibinfo{pages}{8748--8763}.
\newblock


\bibitem[Radford et~al\mbox{.}(2019)]%
        {radford2019language}
\bibfield{author}{\bibinfo{person}{Alec Radford}, \bibinfo{person}{Jeffrey Wu}, \bibinfo{person}{Rewon Child}, \bibinfo{person}{David Luan}, \bibinfo{person}{Dario Amodei}, \bibinfo{person}{Ilya Sutskever}, {et~al\mbox{.}}} \bibinfo{year}{2019}\natexlab{}.
\newblock \showarticletitle{Language models are unsupervised multitask learners}.
\newblock \bibinfo{journal}{\emph{OpenAI blog}} \bibinfo{volume}{1}, \bibinfo{number}{8} (\bibinfo{year}{2019}), \bibinfo{pages}{9}.
\newblock


\bibitem[Ramaswamy et~al\mbox{.}(2019)]%
        {ramaswamy2019federated}
\bibfield{author}{\bibinfo{person}{Swaroop Ramaswamy}, \bibinfo{person}{Rajiv Mathews}, \bibinfo{person}{Kanishka Rao}, {and} \bibinfo{person}{Fran{\c{c}}oise Beaufays}.} \bibinfo{year}{2019}\natexlab{}.
\newblock \showarticletitle{Federated learning for emoji prediction in a mobile keyboard}.
\newblock \bibinfo{journal}{\emph{arXiv preprint arXiv:1906.04329}} (\bibinfo{year}{2019}).
\newblock


\bibitem[Ramesh et~al\mbox{.}(2022)]%
        {ramesh2022hierarchical}
\bibfield{author}{\bibinfo{person}{Aditya Ramesh}, \bibinfo{person}{Prafulla Dhariwal}, \bibinfo{person}{Alex Nichol}, \bibinfo{person}{Casey Chu}, {and} \bibinfo{person}{Mark Chen}.} \bibinfo{year}{2022}\natexlab{}.
\newblock \showarticletitle{Hierarchical text-conditional image generation with clip latents}.
\newblock \bibinfo{journal}{\emph{arXiv preprint arXiv:2204.06125}} \bibinfo{volume}{1}, \bibinfo{number}{2} (\bibinfo{year}{2022}), \bibinfo{pages}{3}.
\newblock


\bibitem[Razavi et~al\mbox{.}(2019)]%
        {razavi2019generating}
\bibfield{author}{\bibinfo{person}{Ali Razavi}, \bibinfo{person}{Aaron Van~den Oord}, {and} \bibinfo{person}{Oriol Vinyals}.} \bibinfo{year}{2019}\natexlab{}.
\newblock \showarticletitle{Generating diverse high-fidelity images with vq-vae-2}.
\newblock \bibinfo{journal}{\emph{Advances in neural information processing systems}}  \bibinfo{volume}{32} (\bibinfo{year}{2019}).
\newblock


\bibitem[Reddi et~al\mbox{.}(2020)]%
        {reddi2020adaptive}
\bibfield{author}{\bibinfo{person}{Sashank Reddi}, \bibinfo{person}{Zachary Charles}, \bibinfo{person}{Manzil Zaheer}, \bibinfo{person}{Zachary Garrett}, \bibinfo{person}{Keith Rush}, \bibinfo{person}{Jakub Kone{\v{c}}n{\`y}}, \bibinfo{person}{Sanjiv Kumar}, {and} \bibinfo{person}{H~Brendan McMahan}.} \bibinfo{year}{2020}\natexlab{}.
\newblock \showarticletitle{Adaptive federated optimization}.
\newblock \bibinfo{journal}{\emph{arXiv preprint arXiv:2003.00295}} (\bibinfo{year}{2020}).
\newblock


\bibitem[Rombach et~al\mbox{.}(2022)]%
        {rombach2022high}
\bibfield{author}{\bibinfo{person}{Robin Rombach}, \bibinfo{person}{Andreas Blattmann}, \bibinfo{person}{Dominik Lorenz}, \bibinfo{person}{Patrick Esser}, {and} \bibinfo{person}{Bj{\"o}rn Ommer}.} \bibinfo{year}{2022}\natexlab{}.
\newblock \showarticletitle{High-resolution image synthesis with latent diffusion models}. In \bibinfo{booktitle}{\emph{Proceedings of the IEEE/CVF conference on computer vision and pattern recognition}}. \bibinfo{pages}{10684--10695}.
\newblock


\bibitem[Saharia et~al\mbox{.}(2022a)]%
        {saharia2022palette}
\bibfield{author}{\bibinfo{person}{Chitwan Saharia}, \bibinfo{person}{William Chan}, \bibinfo{person}{Huiwen Chang}, \bibinfo{person}{Chris Lee}, \bibinfo{person}{Jonathan Ho}, \bibinfo{person}{Tim Salimans}, \bibinfo{person}{David Fleet}, {and} \bibinfo{person}{Mohammad Norouzi}.} \bibinfo{year}{2022}\natexlab{a}.
\newblock \showarticletitle{Palette: Image-to-image diffusion models}. In \bibinfo{booktitle}{\emph{ACM SIGGRAPH 2022 Conference Proceedings}}. \bibinfo{pages}{1--10}.
\newblock


\bibitem[Saharia et~al\mbox{.}(2022b)]%
        {saharia2022photorealistic}
\bibfield{author}{\bibinfo{person}{Chitwan Saharia}, \bibinfo{person}{William Chan}, \bibinfo{person}{Saurabh Saxena}, \bibinfo{person}{Lala Li}, \bibinfo{person}{Jay Whang}, \bibinfo{person}{Emily~L Denton}, \bibinfo{person}{Kamyar Ghasemipour}, \bibinfo{person}{Raphael Gontijo~Lopes}, \bibinfo{person}{Burcu Karagol~Ayan}, \bibinfo{person}{Tim Salimans}, {et~al\mbox{.}}} \bibinfo{year}{2022}\natexlab{b}.
\newblock \showarticletitle{Photorealistic text-to-image diffusion models with deep language understanding}.
\newblock \bibinfo{journal}{\emph{Advances in Neural Information Processing Systems}}  \bibinfo{volume}{35} (\bibinfo{year}{2022}), \bibinfo{pages}{36479--36494}.
\newblock


\bibitem[Saharia et~al\mbox{.}(2022c)]%
        {saharia2022image}
\bibfield{author}{\bibinfo{person}{Chitwan Saharia}, \bibinfo{person}{Jonathan Ho}, \bibinfo{person}{William Chan}, \bibinfo{person}{Tim Salimans}, \bibinfo{person}{David~J Fleet}, {and} \bibinfo{person}{Mohammad Norouzi}.} \bibinfo{year}{2022}\natexlab{c}.
\newblock \showarticletitle{Image super-resolution via iterative refinement}.
\newblock \bibinfo{journal}{\emph{IEEE Transactions on Pattern Analysis and Machine Intelligence}} \bibinfo{volume}{45}, \bibinfo{number}{4} (\bibinfo{year}{2022}), \bibinfo{pages}{4713--4726}.
\newblock


\bibitem[Saputra et~al\mbox{.}(2019)]%
        {saputra2019energy}
\bibfield{author}{\bibinfo{person}{Yuris~Mulya Saputra}, \bibinfo{person}{Dinh~Thai Hoang}, \bibinfo{person}{Diep~N Nguyen}, \bibinfo{person}{Eryk Dutkiewicz}, \bibinfo{person}{Markus~Dominik Mueck}, {and} \bibinfo{person}{Srikathyayani Srikanteswara}.} \bibinfo{year}{2019}\natexlab{}.
\newblock \showarticletitle{Energy demand prediction with federated learning for electric vehicle networks}. In \bibinfo{booktitle}{\emph{2019 IEEE global communications conference (GLOBECOM)}}. IEEE, \bibinfo{pages}{1--6}.
\newblock


\bibitem[Sariyildiz et~al\mbox{.}(2022)]%
        {sariyildiz2022fake}
\bibfield{author}{\bibinfo{person}{Mert~Bulent Sariyildiz}, \bibinfo{person}{Karteek Alahari}, \bibinfo{person}{Diane Larlus}, {and} \bibinfo{person}{Yannis Kalantidis}.} \bibinfo{year}{2022}\natexlab{}.
\newblock \showarticletitle{Fake it till you make it: Learning (s) from a synthetic ImageNet clone}.
\newblock \bibinfo{journal}{\emph{arXiv preprint arXiv:2212.08420}} (\bibinfo{year}{2022}).
\newblock


\bibitem[Sattler et~al\mbox{.}(2021)]%
        {sattler2021fedaux}
\bibfield{author}{\bibinfo{person}{Felix Sattler}, \bibinfo{person}{Tim Korjakow}, \bibinfo{person}{Roman Rischke}, {and} \bibinfo{person}{Wojciech Samek}.} \bibinfo{year}{2021}\natexlab{}.
\newblock \showarticletitle{Fedaux: Leveraging unlabeled auxiliary data in federated learning}.
\newblock \bibinfo{journal}{\emph{IEEE Transactions on Neural Networks and Learning Systems}} (\bibinfo{year}{2021}).
\newblock


\bibitem[Sauer et~al\mbox{.}(2023)]%
        {sauer2023adversarial}
\bibfield{author}{\bibinfo{person}{Axel Sauer}, \bibinfo{person}{Dominik Lorenz}, \bibinfo{person}{Andreas Blattmann}, {and} \bibinfo{person}{Robin Rombach}.} \bibinfo{year}{2023}\natexlab{}.
\newblock \showarticletitle{Adversarial Diffusion Distillation}.
\newblock \bibinfo{journal}{\emph{arXiv preprint arXiv:2311.17042}} (\bibinfo{year}{2023}).
\newblock


\bibitem[Shin et~al\mbox{.}(2020)]%
        {shin2020xor}
\bibfield{author}{\bibinfo{person}{MyungJae Shin}, \bibinfo{person}{Chihoon Hwang}, \bibinfo{person}{Joongheon Kim}, \bibinfo{person}{Jihong Park}, \bibinfo{person}{Mehdi Bennis}, {and} \bibinfo{person}{Seong-Lyun Kim}.} \bibinfo{year}{2020}\natexlab{}.
\newblock \showarticletitle{Xor mixup: Privacy-preserving data augmentation for one-shot federated learning}.
\newblock \bibinfo{journal}{\emph{arXiv preprint arXiv:2006.05148}} (\bibinfo{year}{2020}).
\newblock


\bibitem[Shipard et~al\mbox{.}(2023)]%
        {shipard2023diversity}
\bibfield{author}{\bibinfo{person}{Jordan Shipard}, \bibinfo{person}{Arnold Wiliem}, \bibinfo{person}{Kien~Nguyen Thanh}, \bibinfo{person}{Wei Xiang}, {and} \bibinfo{person}{Clinton Fookes}.} \bibinfo{year}{2023}\natexlab{}.
\newblock \showarticletitle{Diversity is Definitely Needed: Improving Model-Agnostic Zero-shot Classification via Stable Diffusion}. In \bibinfo{booktitle}{\emph{Proceedings of the IEEE/CVF Conference on Computer Vision and Pattern Recognition}}. \bibinfo{pages}{769--778}.
\newblock


\bibitem[Singh and Jaggi(2020)]%
        {singh2020model}
\bibfield{author}{\bibinfo{person}{Sidak~Pal Singh} {and} \bibinfo{person}{Martin Jaggi}.} \bibinfo{year}{2020}\natexlab{}.
\newblock \showarticletitle{Model fusion via optimal transport}.
\newblock \bibinfo{journal}{\emph{Advances in Neural Information Processing Systems}}  \bibinfo{volume}{33} (\bibinfo{year}{2020}), \bibinfo{pages}{22045--22055}.
\newblock


\bibitem[Stypulkowski et~al\mbox{.}(2024)]%
        {stypulkowski2024diffused}
\bibfield{author}{\bibinfo{person}{Michal Stypulkowski}, \bibinfo{person}{Konstantinos Vougioukas}, \bibinfo{person}{Sen He}, \bibinfo{person}{Maciej Zieba}, \bibinfo{person}{Stavros Petridis}, {and} \bibinfo{person}{Maja Pantic}.} \bibinfo{year}{2024}\natexlab{}.
\newblock \showarticletitle{Diffused heads: Diffusion models beat GANs on talking-face generation}. In \bibinfo{booktitle}{\emph{Proceedings of the IEEE/CVF Winter Conference on Applications of Computer Vision}}. \bibinfo{pages}{5091--5100}.
\newblock


\bibitem[Trabucco et~al\mbox{.}(2023)]%
        {trabucco2023effective}
\bibfield{author}{\bibinfo{person}{Brandon Trabucco}, \bibinfo{person}{Kyle Doherty}, \bibinfo{person}{Max Gurinas}, {and} \bibinfo{person}{Ruslan Salakhutdinov}.} \bibinfo{year}{2023}\natexlab{}.
\newblock \showarticletitle{Effective data augmentation with diffusion models}.
\newblock \bibinfo{journal}{\emph{arXiv preprint arXiv:2302.07944}} (\bibinfo{year}{2023}).
\newblock


\bibitem[Vahidian et~al\mbox{.}(2023)]%
        {vahidian2023efficient}
\bibfield{author}{\bibinfo{person}{Saeed Vahidian}, \bibinfo{person}{Mahdi Morafah}, \bibinfo{person}{Weijia Wang}, \bibinfo{person}{Vyacheslav Kungurtsev}, \bibinfo{person}{Chen Chen}, \bibinfo{person}{Mubarak Shah}, {and} \bibinfo{person}{Bill Lin}.} \bibinfo{year}{2023}\natexlab{}.
\newblock \showarticletitle{Efficient distribution similarity identification in clustered federated learning via principal angles between client data subspaces}. In \bibinfo{booktitle}{\emph{Proceedings of the AAAI Conference on Artificial Intelligence}}, Vol.~\bibinfo{volume}{37}. \bibinfo{pages}{10043--10052}.
\newblock


\bibitem[von Platen et~al\mbox{.}(2022a)]%
        {von-platen-etal-2022-diffusers}
\bibfield{author}{\bibinfo{person}{Patrick von Platen}, \bibinfo{person}{Suraj Patil}, \bibinfo{person}{Anton Lozhkov}, \bibinfo{person}{Pedro Cuenca}, \bibinfo{person}{Nathan Lambert}, \bibinfo{person}{Kashif Rasul}, \bibinfo{person}{Mishig Davaadorj}, {and} \bibinfo{person}{Thomas Wolf}.} \bibinfo{year}{2022}\natexlab{a}.
\newblock \bibinfo{title}{Diffusers: State-of-the-art diffusion models}.
\newblock \bibinfo{howpublished}{\url{https://github.com/huggingface/diffusers}}.
\newblock


\bibitem[von Platen et~al\mbox{.}(2022b)]%
        {von2022diffusers}
\bibfield{author}{\bibinfo{person}{Patrick von Platen}, \bibinfo{person}{Suraj Patil}, \bibinfo{person}{Anton Lozhkov}, \bibinfo{person}{Pedro Cuenca}, \bibinfo{person}{Nathan Lambert}, \bibinfo{person}{Kashif Rasul}, \bibinfo{person}{Mishig Davaadorj}, {and} \bibinfo{person}{Thomas Wolf}.} \bibinfo{year}{2022}\natexlab{b}.
\newblock \bibinfo{title}{Diffusers: State-of-the-art diffusion models}.
\newblock
\newblock


\bibitem[Wang et~al\mbox{.}(2020b)]%
        {wang2020federated}
\bibfield{author}{\bibinfo{person}{Hongyi Wang}, \bibinfo{person}{Mikhail Yurochkin}, \bibinfo{person}{Yuekai Sun}, \bibinfo{person}{Dimitris Papailiopoulos}, {and} \bibinfo{person}{Yasaman Khazaeni}.} \bibinfo{year}{2020}\natexlab{b}.
\newblock \showarticletitle{Federated learning with matched averaging}.
\newblock \bibinfo{journal}{\emph{arXiv preprint arXiv:2002.06440}} (\bibinfo{year}{2020}).
\newblock


\bibitem[Wang et~al\mbox{.}(2020a)]%
        {wang2020tackling}
\bibfield{author}{\bibinfo{person}{Jianyu Wang}, \bibinfo{person}{Qinghua Liu}, \bibinfo{person}{Hao Liang}, \bibinfo{person}{Gauri Joshi}, {and} \bibinfo{person}{H~Vincent Poor}.} \bibinfo{year}{2020}\natexlab{a}.
\newblock \showarticletitle{Tackling the objective inconsistency problem in heterogeneous federated optimization}.
\newblock \bibinfo{journal}{\emph{Advances in neural information processing systems}}  \bibinfo{volume}{33} (\bibinfo{year}{2020}), \bibinfo{pages}{7611--7623}.
\newblock


\bibitem[Wang et~al\mbox{.}(2018)]%
        {wang2018dataset}
\bibfield{author}{\bibinfo{person}{Tongzhou Wang}, \bibinfo{person}{Jun-Yan Zhu}, \bibinfo{person}{Antonio Torralba}, {and} \bibinfo{person}{Alexei~A Efros}.} \bibinfo{year}{2018}\natexlab{}.
\newblock \showarticletitle{Dataset distillation}.
\newblock \bibinfo{journal}{\emph{arXiv preprint arXiv:1811.10959}} (\bibinfo{year}{2018}).
\newblock


\bibitem[Yang et~al\mbox{.}(2023)]%
        {yang2023diffusion}
\bibfield{author}{\bibinfo{person}{Ling Yang}, \bibinfo{person}{Zhilong Zhang}, \bibinfo{person}{Yang Song}, \bibinfo{person}{Shenda Hong}, \bibinfo{person}{Runsheng Xu}, \bibinfo{person}{Yue Zhao}, \bibinfo{person}{Wentao Zhang}, \bibinfo{person}{Bin Cui}, {and} \bibinfo{person}{Ming-Hsuan Yang}.} \bibinfo{year}{2023}\natexlab{}.
\newblock \showarticletitle{Diffusion models: A comprehensive survey of methods and applications}.
\newblock \bibinfo{journal}{\emph{Comput. Surveys}} \bibinfo{volume}{56}, \bibinfo{number}{4} (\bibinfo{year}{2023}), \bibinfo{pages}{1--39}.
\newblock


\bibitem[Yoon et~al\mbox{.}(2021)]%
        {yoon2021fedmix}
\bibfield{author}{\bibinfo{person}{Tehrim Yoon}, \bibinfo{person}{Sumin Shin}, \bibinfo{person}{Sung~Ju Hwang}, {and} \bibinfo{person}{Eunho Yang}.} \bibinfo{year}{2021}\natexlab{}.
\newblock \showarticletitle{Fedmix: Approximation of mixup under mean augmented federated learning}.
\newblock \bibinfo{journal}{\emph{arXiv preprint arXiv:2107.00233}} (\bibinfo{year}{2021}).
\newblock


\bibitem[Yurochkin et~al\mbox{.}(2019)]%
        {yurochkin2019bayesian}
\bibfield{author}{\bibinfo{person}{Mikhail Yurochkin}, \bibinfo{person}{Mayank Agarwal}, \bibinfo{person}{Soumya Ghosh}, \bibinfo{person}{Kristjan Greenewald}, \bibinfo{person}{Nghia Hoang}, {and} \bibinfo{person}{Yasaman Khazaeni}.} \bibinfo{year}{2019}\natexlab{}.
\newblock \showarticletitle{Bayesian nonparametric federated learning of neural networks}. In \bibinfo{booktitle}{\emph{International conference on machine learning}}. PMLR, \bibinfo{pages}{7252--7261}.
\newblock


\bibitem[Zhang et~al\mbox{.}(2023b)]%
        {zhang2023text}
\bibfield{author}{\bibinfo{person}{Chenshuang Zhang}, \bibinfo{person}{Chaoning Zhang}, \bibinfo{person}{Mengchun Zhang}, {and} \bibinfo{person}{In~So Kweon}.} \bibinfo{year}{2023}\natexlab{b}.
\newblock \showarticletitle{Text-to-image diffusion model in generative ai: A survey}.
\newblock \bibinfo{journal}{\emph{arXiv preprint arXiv:2303.07909}} (\bibinfo{year}{2023}).
\newblock


\bibitem[Zhang et~al\mbox{.}(2023a)]%
        {zhang2023gptfl}
\bibfield{author}{\bibinfo{person}{Tuo Zhang}, \bibinfo{person}{Tiantian Feng}, \bibinfo{person}{Samiul Alam}, \bibinfo{person}{Dimitrios Dimitriadis}, \bibinfo{person}{Mi Zhang}, \bibinfo{person}{Shrikanth~S. Narayanan}, {and} \bibinfo{person}{Salman Avestimehr}.} \bibinfo{year}{2023}\natexlab{a}.
\newblock \bibinfo{title}{GPT-FL: Generative Pre-trained Model-Assisted Federated Learning}.
\newblock
\newblock
\showeprint[arxiv]{2306.02210}~[cs.LG]


\bibitem[Zhao et~al\mbox{.}(2018)]%
        {zhao2018federated}
\bibfield{author}{\bibinfo{person}{Yue Zhao}, \bibinfo{person}{Meng Li}, \bibinfo{person}{Liangzhen Lai}, \bibinfo{person}{Naveen Suda}, \bibinfo{person}{Damon Civin}, {and} \bibinfo{person}{Vikas Chandra}.} \bibinfo{year}{2018}\natexlab{}.
\newblock \showarticletitle{Federated learning with non-iid data}.
\newblock \bibinfo{journal}{\emph{arXiv preprint arXiv:1806.00582}} (\bibinfo{year}{2018}).
\newblock


\bibitem[Zhou et~al\mbox{.}(2020)]%
        {zhou2020distilled}
\bibfield{author}{\bibinfo{person}{Yanlin Zhou}, \bibinfo{person}{George Pu}, \bibinfo{person}{Xiyao Ma}, \bibinfo{person}{Xiaolin Li}, {and} \bibinfo{person}{Dapeng Wu}.} \bibinfo{year}{2020}\natexlab{}.
\newblock \showarticletitle{Distilled one-shot federated learning}.
\newblock \bibinfo{journal}{\emph{arXiv preprint arXiv:2009.07999}} (\bibinfo{year}{2020}).
\newblock


\bibitem[Zhu et~al\mbox{.}(2021b)]%
        {zhu2021federated}
\bibfield{author}{\bibinfo{person}{Hangyu Zhu}, \bibinfo{person}{Jinjin Xu}, \bibinfo{person}{Shiqing Liu}, {and} \bibinfo{person}{Yaochu Jin}.} \bibinfo{year}{2021}\natexlab{b}.
\newblock \showarticletitle{Federated learning on non-IID data: A survey}.
\newblock \bibinfo{journal}{\emph{Neurocomputing}}  \bibinfo{volume}{465} (\bibinfo{year}{2021}), \bibinfo{pages}{371--390}.
\newblock


\bibitem[Zhu et~al\mbox{.}(2021a)]%
        {zhu2021data}
\bibfield{author}{\bibinfo{person}{Zhuangdi Zhu}, \bibinfo{person}{Junyuan Hong}, {and} \bibinfo{person}{Jiayu Zhou}.} \bibinfo{year}{2021}\natexlab{a}.
\newblock \showarticletitle{Data-free knowledge distillation for heterogeneous federated learning}. In \bibinfo{booktitle}{\emph{International conference on machine learning}}. PMLR, \bibinfo{pages}{12878--12889}.
\newblock


\end{thebibliography}

\end{document}